# VLM-SAFE: Vision-Language Model-Guided Safety-Aware Reinforcement Learning with World Models for Autonomous Driving

Yansong Qu †, Zilin Huang †, Zihao Sheng †, Jiancong Chen, Yue Leng, Samuel Labi, Sikai Chen *

***Abstract*—Autonomous driving policy learning with reinforcement learning (RL) is fundamentally limited by low sample efficiency, weak generalization, and a dependence on unsafe online trial-and-error interactions. Although safe RL introduces explicit constraints or costs, existing methods often fail to capture the semantic meaning of safety in real driving scenes, leading to conservative behaviors in simple cases and insufficient risk awareness in complex ones. To address this issue, we propose VLM-SAFE, an offline safe RL framework that follows a human cognitive loop of observe–imagine–evaluate–act. Starting from offline driving data, VLM-SAFE observes traffic scenarios and leverages a vision-language model (VLM) to provide semantic safety signals grounded in scene understanding. A learned world model then imagines future trajectories from the observed context, enabling the agent to reason about possible consequences without interacting with the real environment. Rather than using imagined rollouts solely for return estimation, VLM-SAFE further evaluates these predicted futures with VLM-based safety guidance, explicitly coupling future anticipation with semantic risk assessment. The resulting safety-aware imagined experience is finally used to optimize the policy via actor-critic learning, such that actions are chosen based on both predicted outcomes and their safety implications. By tightly integrating observation, imagination, evaluation, and action into a unified closed loop, VLM-SAFE enables safer and more efficient offline policy learning for autonomous driving. Extensive experiments in simulation show that VLM-SAFE achieves improved safety, stronger robustness under traffic-density shift, and a better safety–performance trade-off than representative baselines.**

## 1. INTRODUCTION

Autonomous driving has made tremendous progress over the past decade, yet modern transportation systems still suffer from persistent inefficiencies and limited capabilities in accident prevention and mobility optimization (Long et al., 2024, 2025). Autonomous driving, with its advanced intelligence and automation (Sheng et al., 2024), offers a promising solution. At its core, autonomous driving aims to perceive (Qu et al., 2025), understand (Zhang et al., 2024), and interact (Z. Huang et al., 2024) with complex dynamic systems in real time. Among various approaches, reinforcement learning (RL) (Fig. 1(a)) offers a principled paradigm for developing intelligent driving agents: by interacting with an environment, an agent learns a policy that maximizes expected long-term returns. However, the very nature of RL, trial-and-error exploration, creates a fundamental tension in safety-critical applications. More interaction can accelerate learning, but it also increases the chance of unsafe behaviors. This motivates a long line of research on enabling effective learning while controlling risk.

A natural direction is to reduce or remove online exploration. Imitation learning (IL) (Fig. 1(b)) and offline RL (Fig. 1(c)) learn from pre-collected expert demonstrations or logged driving data, thereby avoiding unsafe real-time trial-and-error. Yet, offline learning is still not “safe by default”: when deployed, the learned policy may face distribution shift and corner cases that are underrepresented in the dataset, leading to brittle behaviors. In parallel, world-model-based RL (Hafner et al., 2019, 2020, 2023) (Fig. 1(d)) has emerged as a new generation of approaches that improve sample efficiency by learning a predictive model of the environment and optimizing policies through *imagined rollouts* instead of relying solely on costly interaction. By “dreaming” future trajectories, the agent can plan and learn more efficiently and potentially reduce risky exploration. Nevertheless, imagination alone does not guarantee safety, especially in the early stages when the world model is inaccurate, or in complex scenarios where safety depends on subtle contextual cues.

**Corresponding author:* Sikai Chen (e-mail: sikai.chen@wisc.edu). † Equal contribution.

Yansong Qu is with the Lyles School of Civil and Construction Engineering, Purdue University, West Lafayette, 47907, USA (e-mail: qu120@purdu.edu).

Zilin Huang is with the Department of Civil and Environmental Engineering, University of Wisconsin-Madison, Madison, 53706, USA (e-mail: zilin.huang@wisc.edu).

Zihao Sheng is with the Department of Civil and Environmental Engineering, University of Wisconsin-Madison, Madison, 53706, USA (e-mail: zihao.sheng@wisc.edu).

Jiancong Chen is with the Lyles School of Civil and Construction Engineering, Purdue University, West Lafayette, 47907, USA (e-mail: chen5281@purdue.edu).

Yue Leng is with the Google, Sunnyvale, CA 94089, USA (e-mail: cathylengyue@gmail.com).

Samuel Labi is with the Lyles School of Civil and Construction Engineering, Purdue University, West Lafayette, 47907, USA (e-mail: labi@purdue.edu).

Sikai Chen is with the Department of Civil and Environmental Engineering, University of Wisconsin-Madison, Madison, 53706, USA (e-mail: sikai.chen@wisc.edu).

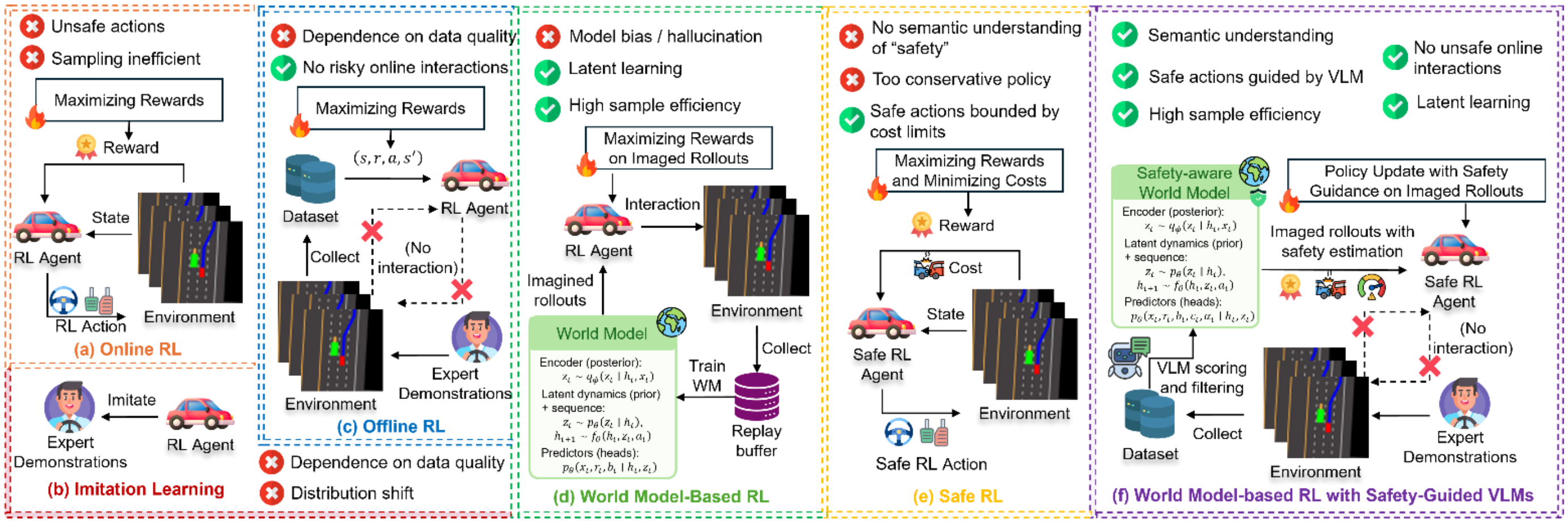

**Fig. 1.** Comparison between the proposed method and related approaches.

To address safety explicitly, safe RL (Fig. 1(e)) formulates the driving problem as a constrained optimization problem, typically introducing a cost signal as a risk surrogate and using principled primal-dual (Lagrangian) updates to balance reward maximization and cost minimization. Recent advances further integrate safe RL with world models, demonstrating strong performance in safety benchmarks. For example, FOSP (Cao et al., 2024) leverages world models to fine-tune an offline safe policy and construct safe policy improvement using value-based or reachability/feasibility guidance.

A key commonality across these safe world-model-based methods is that "safety" is operationalized through task-defined costs or constraint proxies. Consequently, policy improvement is controlled by weights or coefficients derived from these signals: Lagrangian multipliers reweight reward versus cost, feasibility/reachability modules influence safe action selection, and value-based distributions shape conservative updates. These mechanisms are effective; however, autonomous driving safety is often semantic and context-dependent: critical hazards may be indicated by visual cues and background context that are difficult to exhaustively encode into a sparse cost function.

This motivates a human-inspired view of safe decision-making, illustrated in Fig. 2. In the human cognitive loop (left panel), safety is rarely judged only after a bad outcome happens; instead, people observe a cue, imagine plausible consequences, semantically evaluate the risk, and then act proactively. For example, a father notices that cooking is producing unusually heavy smoke. He immediately imagines what may happen next: the smoke could trigger the fire alarm, the alarm could wake up the baby, and the situation could become risky and disruptive. Based on this semantic chain of consequences, he decides to turn on the range hood before the alarm is triggered. Inspired by this everyday example, the AI cognitive loop for autonomous driving (right panel) follows the same principle: **observe** (visual state), **imagine** (world-model rollouts), **semantically evaluate risk** (VLM-grounded safety estimation), and **act safely** (Lagrangian-constrained policy optimization with semantic-gated updates). Consider a vehicle approaching a stopped school bus with an extended STOP sign while children are crossing. The key is not to wait for a collision or an explicit penalty to react; rather, the agent imagines "what will happen" in the near future, evaluates the situation as high-risk, and then brake or stop early in a proportionate manner. These examples together suggest that robust autonomy requires context-aware semantic safety understanding, a mechanism that can translate visual cues into risk-aware decisions before catastrophic events occur.

In this work, we propose VLM-SAFE, an offline world-model-based safe RL framework (Fig. 1(f)) that instantiates the AI cognitive loop in Fig. 2 and explicitly equips world-model-based RL with semantic safety. Concretely, we treat a pretrained VLM as a semantic safety critic during offline dataset curation: given a driving observation, the VLM outputs a continuous safety score $\alpha$ with chain-of-thought reasoning that reflects semantic risk grounded in the visual context. We then train a safety-aware world model that not only predicts future observations and rewards, but also learns to predict cost- and safety-related quantities along imagined rollouts. This enables the agent to **"imagine what will happen" together with predicted safety assessments**, matching the "Imagine→Evaluate" steps in the cognitive loop without requiring online VLM calls during policy learning.

Importantly, VLM-SAFE's key novelty lies in how semantic safety grounds policy improvement. Built upon behavior-regularized (KL-constrained) Lagrangian safe RL, the optimal policy update takes an exponential-tilting form where a non-negative weighting term controls how actions are reweighted during improvement. In VLM-SAFE, this weighting is instantiated as a state-level semantic safety gate parameterized by $\alpha$. Intuitively, when $\alpha$ is high, the state is judged semantically safe and reward-oriented improvement is allowed to proceed more aggressively; when $\alpha$ is low, reward-driven updates are automatically suppressed, while the Lagrangian mechanism continues to enforce explicit risk control through the cost objective. In this way, we preserve principled constraint handling and simultaneously inject a semantic layer that makes world-model-based RL safety-aware in a context-dependent and transferable manner. The main contributions of this work are as follows:

(1) We propose VLM-SAFE, an offline safe world-model RL framework that trains on imagined rollouts and explicitly predicts safety-related quantities, improving safety and data efficiency without unsafe online exploration.

(2) We formulate safe autonomous driving policy learning from a human cognitive-loop perspective and propose an observe–imagine–evaluate–act paradigm, in which offline observations provide context, the world model imagines future trajectories, VLM-based safety guidance evaluates the imagined outcomes, and the policy acts under safety-aware optimization.

(3) We introduce a VLM-grounded semantic safety mechanism that provides continuous safety supervision and instantiates a state-level gate to modulate policy improvement strength across contexts, complementing a Lagrangian safe RL backbone.

(4) We demonstrate in simulation that semantic safety grounding improves safety and robustness, particularly under traffic-density shift, while yielding a better safety-performance trade-off than representative baselines.

The remainder of this paper is organized as follows. Section 2 reviews related work. Section 3 introduces preliminaries. Section 4 presents the methodology. Section 5 reports experimental results, and Section 6 concludes.

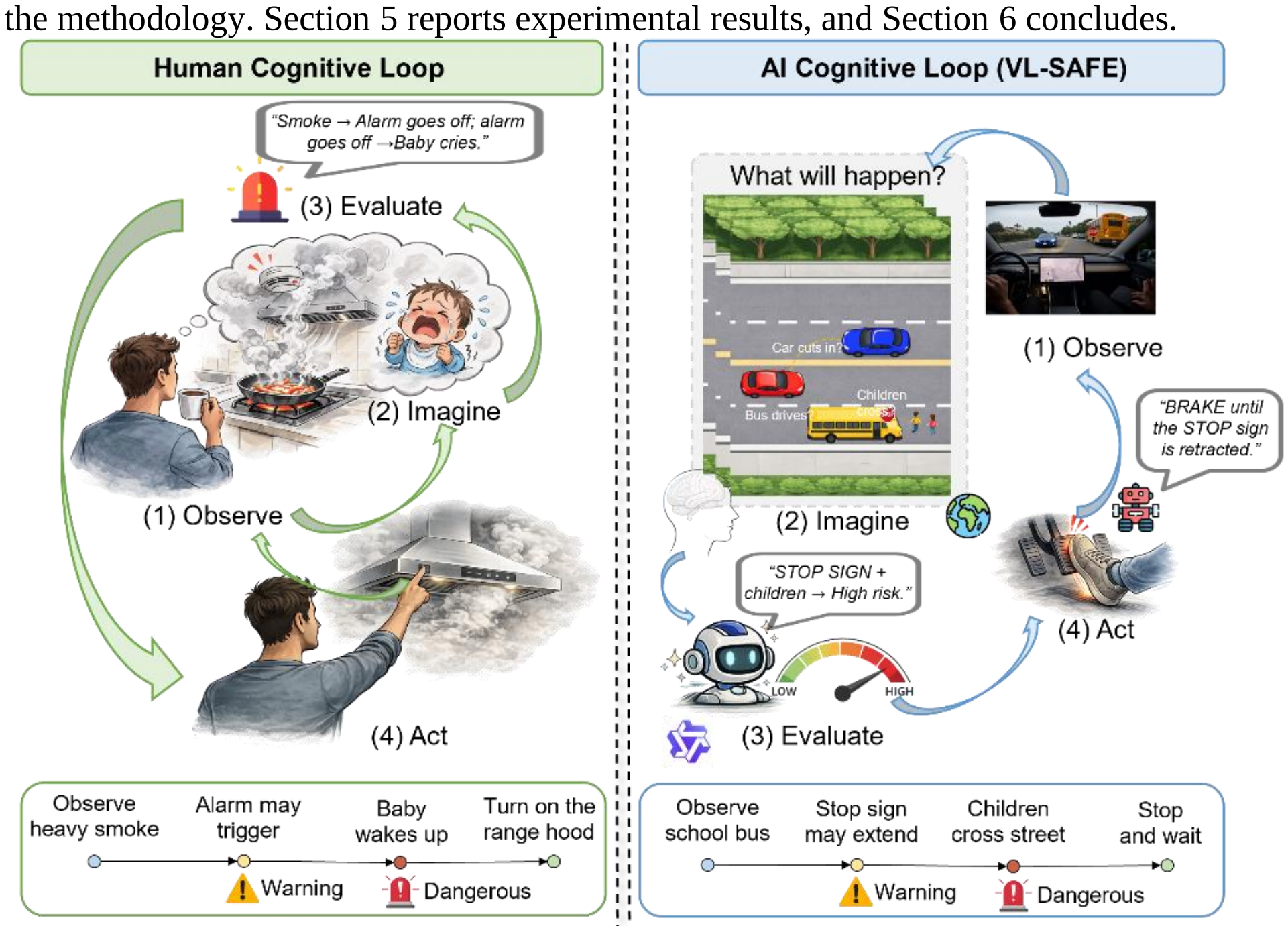

**Fig. 2.** Illustration of the human cognitive loop and the corresponding AI cognitive loop in VLM-SAFE, both following an observe–imagine–evaluate–act process.

## 2. Related Work

### *2.1 Classical online RL-based autonomous driving*

RL offers a principled framework for sequential decision-making and has gained significant traction in autonomous systems research. Despite its promise, RL faces several limitations in practical deployment. Generalization remains a key challenge, as agents trained in simulation often fail in other unseen environments (Wu et al., 2023). Additionally, standard RL algorithms do not explicitly consider safety, potentially leading to risky or catastrophic behavior during both training and execution. By leveraging a world model and VLM-based safety guidance in an offline setting, our proposed method is designed to alleviate these shortcomings.

### *2.2 Offline RL*

By decoupling learning from real-time interactions, offline RL enables the use of previously collected expert demonstrations to develop safety-oriented policies (Levine et al., 2020). Previous researchers (Koirala et al., 2024) have proposed a novel offline RL method that learns a conservative (safe) policy using latent safety constraints and optimizes cumulative rewards under these constraints via advantage-weighted learning in the latent space. However, offline RL policies trained on limited or suboptimal datasets often suffer from distributional shift, making them prone to selecting out-of-distribution actions (Kostrikov et al., 2021; Kumar et al., 2020). Furthermore, safety signals (e.g., risk annotations, constraint violations) are often sparse or unavailable in the offline dataset (Rigter et al., 2022). Moreover, offline RL suffers from limited generalization due to fixed datasets. In contrast, our method leverages a world model to generate imagined rollouts, thereby improving data diversity and sample efficiency.

### *2.3 World models*

World models learn to simulate environmental dynamics from data, enabling agents to plan or imagine rollouts internally (Hafner et al., 2019, 2020, 2023). They help decouple policy learning from direct environment interaction by simulating future trajectories, thus improving sample efficiency, particularly in safety-critical or data-scarce domains such as autonomous driving (Li et al., 2024; X. Wang et al., 2024; Y. Wang et al., 2024). However, a limitation of standard world models is that they may produce inaccurate or unrealistic rollouts, particularly during the early stages of training. This can lead to unsafe actions, as the policy is optimized based on flawed imagined trajectories. To mitigate this issue, our method introduces a safety-aware world model that performs semantic safety evaluation on each imagined state. By identifying risky trajectories and guiding the policy away from them, this mechanism promotes safer policy learning.

### *2.4 Safe RL*

Safe RL aims to optimize agent behavior while satisfying safety constraints throughout training and deployment. Constrained optimization approaches solve the problem by incorporating Lagrangian multipliers or dual gradient methods (Hu et al., 2024). Risk-sensitive approaches formulate objectives that consider conditional value-at-risk, worst-case return, or variance-based measures (W. Huang et al., 2024). Other lines of work leverage control-theoretic techniques (Zhou et al., 2024) or Lyapunov-based constraints (Pan et al., 2024), to ensure safety in continuous control systems. These methods are effective but lack semantic understanding of safety and often overlook driving context, leading to biased (overly aggressive or overly conservative) policies. In contrast, we leverage a VLM to derive safety scores based on scene semantics, enabling more context-aware and balanced safe policy learning.

## 3. PRELIMINARIES

### *3.1 Markov Decision Processes*

A Markov Decision Process (MDP) is defined as a tuple $\mathcal{M} = (\mathcal{S}, \mathcal{A}, \mathcal{T}, \mathcal{R}, \mu, \gamma)$, where $\mathcal{S}$ denotes the state space, $\mathcal{A}$ denotes the action space, $\mathcal{T}: \mathcal{S} \times \mathcal{A} \times \mathcal{S} \rightarrow [0,1]$ denotes the transition probability function, $\mathcal{R}: \mathcal{S} \times \mathcal{A} \rightarrow \mathbb{R}$ denotes reward function, $\mu(\cdot): \mathcal{S} \rightarrow [0,1]$ denotes the initial state distribution, and $\gamma \in [0,1]$ denotes the discount factor. At each timestep, the agent observes state $s \in \mathcal{S}$ and selects an action $a \in \mathcal{A}$. The environment transitions to the next state $s'$ according to the transition function $\mathcal{T}(s'|s,a)$. The agent then receives an immediate reward $R(s,a)$. For trajectory-based formulations, a trajectory can be written as $\tau = (s_0, a_0, r_0, s_1, a_1, r_1, \dots)$, and its return is defined as the discounted sum of rewards, $g(\tau; R) = \sum_{t=0}^{T} \gamma^t R(s_t, a_t)$. The agent's goal is to learn a policy $\pi(a \mid s)$ that maximizes the expected return $J(\pi) = \mathbb{E}_\pi[g(\tau(\pi); R)]$. In practice, the objective can also be expressed through the state-value function $V(s) = \mathbb{E}_{\tau\sim\pi_\psi}[\sum_{t=0}^{T} \gamma^t R(s_t, a_t)]$ or action-state value function $\mathrm{Q}(s,a) = \mathbb{E}_{\tau\sim\pi_\psi}[\sum_{t=0}^{T} \gamma^t R(s_t, a_t)]$.

### *3.2 Offline RL*

In contrast to online RL methods, offline RL uses a fixed dataset to learn an optimal policy without further interacting with the environment. Unlike imitation learning (IL), offline RL does not directly mimic expert behavior and can therefore infer actions that improve upon those contained in the dataset. Offline RL is effective; however, its generalization is limited by the fixed dataset. In contrast, our method incorporates a world model to generate imagined rollouts, thereby enriching the effective training distribution and improving sample efficiency. Offline RL is typically trained by minimizing a temporal-difference (TD) error of the form:

$$L_{TD}(\theta) = \mathbb{E}_{(s,a,s')\sim\mathcal{D}}\left[\left(r(s,a) + \gamma \max_{a'} Q_{\hat{\theta}}(s', a') - Q_\theta(s,a)\right)^2\right] \tag{1}$$

where $\mathcal{D}$ denotes the offline dataset, $Q_\theta(s,a)$ denotes the Q-function parameterized by $\theta$, and $Q_{\hat{\theta}}(s,a)$ denotes the target network with softly updated parameters. A major limitation of offline RL is Q-value overestimation, primarily resulting from extrapolation errors when evaluating actions that are not well-supported by the dataset.

### *3.3 World-model-based safe RL*

Safe RL is formulated as a constrained MDP $\mathcal{M} = (\mathcal{S}, \mathcal{A}, \mathcal{T}, \mathcal{R}, \mathcal{C}, \mu, \gamma)$, where $\mathcal{C}$ denotes a set of cost functions $c: \mathcal{S} \times \mathcal{A} \rightarrow [0, c_{\max}]$. Given a policy $\pi_\psi$, the expected reward and expected cost objectives of safe RL are defined as:

$$J^{\mathcal{R}}(\pi_\psi) = \mathbb{E}_{\boldsymbol{a}_t\sim\pi_\psi, \boldsymbol{s}_{t+1}\sim p_\theta, \boldsymbol{s}_0\sim\mu}\left[\sum_{t=0}^{T} r_t\right] \tag{2}$$

$$J_i^{\mathcal{C}}(\pi_\psi) = \mathbb{E}_{\boldsymbol{a}_t\sim\pi_\psi, \boldsymbol{s}_{t+1}\sim p_\theta, \boldsymbol{s}_0\sim\mu}\left[\sum_{t=0}^{T} c_t^i\right] \leqslant d^i, \forall i \in \{1, \dots, C\} \tag{3}$$

where $c_t^i$ denotes the $i$-th safety constraint, which the policy $\pi_\psi$ is required to satisfy, and $d^i$ denotes the corresponding cost threshold. The safe RL problem can thus be written as:

$$\max_{\psi} J^{\mathcal{R}}(\pi_\psi) \ \text{ s.t. } J_i^{\mathcal{C}}(\pi_\psi) \leqslant d^i, \forall i \in \{1, \dots, C\} \tag{4}$$

Under a model-based setting, generated rollouts from a learned dynamics model can be used to train the policy with predicted rewards and costs. By introducing Lagrange multipliers, the safe RL problem can be reformulated as the following min-max objective:

$$\min_{\lambda \geq 0} \max_{\psi} \left[ J^R(\pi_\psi) - \sum_{i=1}^{C} \lambda_i \left( J_i^C(\pi_\psi) - d^i \right) \right] \tag{5}$$

where the inner maximization optimizes the reward objective while penalizing constraint violations through the Lagrange multipliers $\lambda_i$, and the outer minimization adjusts $\lambda_i$ to enforce the constraints. However, safe RL only constrains the cost under a threshold, not considering the entire driving context. Instead, our proposed VLM-as-safety-guidance paradigm is designed to be capable of semantically understanding the driving context and infer safety scores for each state.

### *3.4 Problem Statement*

We formulate the task of training an agent as a constrained MDP, following (Cao et al., 2024). The agent's objective is to learn an optimal safe policy $\pi_\psi: \mathcal{S} \to \mathcal{A}$ that maximizes the expected cumulative reward while reducing violations of safety constraints. A key challenge is how to evaluate the risk of the current state. Our key idea is to leverage the semantic understanding capability of pretrained VLMs to estimate a safety probability for each state. These safety probabilities are used by the world model as supervision targets to predict the safety of future rollouts, which are then used for policy learning. Under this setting, we augment the constrained MDP as $\mathcal{M} = (\mathcal{S}, \mathcal{A}, \mathcal{T}, \mathcal{R}, \mathcal{C}, \mathcal{A}^{\mathcal{S}}, \mu, \gamma)$, where $\alpha \in \mathcal{A}^{\mathcal{S}}$ denotes the state-wise safety probability estimated by the VLM.

## 4. METHODOLOGY

As shown in Fig. 2, VLM-SAFE is designed as a concrete instantiation of the human-inspired observe–imagine–evaluate–act cognitive loop for offline safe autonomous driving. Specifically, the first training stage (left panel in Fig. 3) corresponds to the observe and initial evaluate steps: given offline driving data, a pre-trained VLM provides safety estimates for each observed state in the dataset (Section 4.1). The second training stage (right panel in Fig. 3) realizes the imagine–evaluate–act process: a safety-aware world model is trained to generate imagined rollouts together with predicted safety estimates, based on which actor–critic learning is performed under safety guidance to optimize the driving policy (Sections 4.2–4.4). In this way, VLM-SAFE forms a closed loop that connects offline observation, future imagination, semantic safety evaluation, and policy improvement.

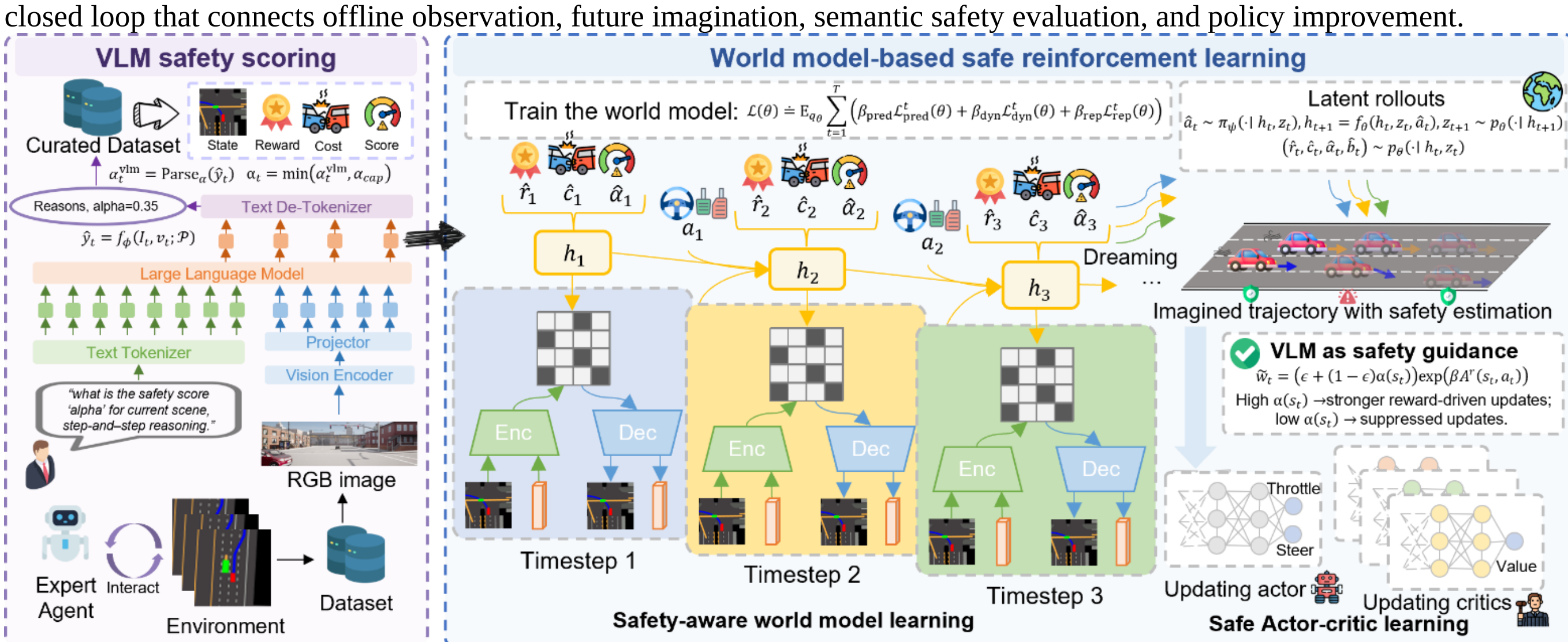

**Fig. 3.** Overall framework of VLM-SAFE.

### *4.1 VLM safety scoring*

Given a state $s_t \in \mathcal{S}$, we assign a continuous semantic safety score $\alpha_t \in [0,1]$ to represent the safety level of the current driving situation. In our setting, each state is represented by an observation tuple $s_t = (I_t, v_t)$, where $I_t$ is the front-facing camera image and $v_t$ is the ego speed. We use a pretrained VLM as a semantic evaluator to generate an offline safety label for each state, denoted by $\alpha_t^{\text{vlm}}$.

Specifically, we query the VLM with the structured prompt shown in Fig. 4, which enforces an ordered reasoning procedure and produces a machine-readable output. Let $f_\phi$ denote the pretrained VLM with parameters $\phi$, and let $\mathcal{P}$ denote the prompt template. The VLM returns a JSON object containing a textual rationale and a numeric field "alpha". We formalize the extraction as

$$\hat{y}_t = f_\phi(I_t, v_t; \mathcal{P}), \alpha_t^{\text{vlm}} = \text{Parse}_\alpha(\hat{y}_t), \tag{6}$$

where $\mathrm{Parse}_{\alpha}(\cdot)$ extracts the scalar value from the JSON field "alpha", and $\alpha_t^{\mathrm{vlm}} \in [0,1]$.

The prompt in Fig. 4 provides an anchored continuous risk scale (from 1.0, “ideal,” to 0.0-0.1, “critical”) and instructs the VLM to interpolate smoothly between anchors based on perceived risk, thereby treating safety scoring as continuous regression rather than discrete classification. In addition to dynamic agent and context scanning, the prompt explicitly includes a safety-critical "occlusion check" step that encourages conservative scoring when the forward view is heavily blocked at close range. We use only the final numeric output $\alpha_t^{\mathrm{vlm}}$ as the label, while the rationale text is retained solely for debugging and auditing.

Task: Autonomous Driving Safety Assessment (Continuous Regression).
Input: Front-facing camera image + Ego Speed.
Ego Speed: Moving at Slow speed (3.9 m/s).
Act as an experienced driver. Predict a precise continuous safety score 'alpha' (0.00 to 1.00).
Step-by-Step Reasoning (You MUST follow this order in your output):

1. **CRITICAL OCCLUSION CHECK**: - Is the view blocked by a **Car Rear**, **Wall**, or **Large Truck** (< 2m)? - If YES -> STOP analysis, Score = 0.05.
2. **SCENE SCAN**: - List dynamic agents (Pedestrians, Vehicles) and static context (Lights, Lane width).
3. **DETERMINE SCORE (Interpolate smoothly)**:
   - **1.0 (Ideal)**: Empty highway/desert road. Zero objects.
   - **0.9 (Very Safe)**: Open road, but urban environment (buildings/trees visible).
   - **0.8 (Safe)**: Path clear, but simple complexity (e.g., parked cars on side).
   - **0.7 (Standard)**: Typical city traffic. Car ahead at safe distance, or green light.
   - **0.6 (Alert)**: Scene is busy. Multiple cars, or narrow lane.
   - **0.5 (Caution)**: Potential conflict. Pedestrian near curb, or Lead Car brake lights.
   - **0.4 (Unsafe)**: Yellow light, or Lead Car slowing down fast.
   - **0.3 (Warning)**: Red light, or Lead Car close (< 10m).
   - **0.2 (Danger)**: Cyclist/Pedestrian in ego lane.
   - **0.0 - 0.1 (Critical)**: Crash imminent or View Occluded.

**Scoring Logic**: Use the full range. Interpolate between anchors based on risk.
Output JSON: {{"reason": "Step 1: [Occlusion Check Result]... Step 2: [List Objects]... Step 3: [Reasoning for Score]...", "alpha": <float>}}

**Fig. 4.** Text prompt for initial offline safety evaluation.

Since we rely on a front-facing RGB view, some risk factors may be only partially observable (e.g., events originating outside the camera's field of view). To reduce overly optimistic labels in such cases, we apply a lightweight conservative filtering step during offline dataset curation using an environment-provided high-risk indicator $c_t \in [0,1]$. Specifically, when the state is flagged as high-risk ($c_t > \tau_c$), we cap the final safety label by a conservative bound:

$$\alpha_t = \min\left(\alpha_t^{\mathrm{vlm}}, \alpha_{cap}\right), \tag{7}$$

Otherwise, we keep $\alpha_t = \alpha_t^{\mathrm{vlm}}$. Here $\tau_c$ and $\alpha_{\mathrm{cap}}$ are fixed hyperparameters. This filtering is performed only once for offline label generation and is not required at inference time.

Overall, the VLM is used only during offline dataset curation to generate $\alpha_t$ for training. Subsequent model learning and policy optimization do not require invoking the VLM online; instead, the learned components introduced in the following sections rely on predicted safety-related quantities, which substantially reduces computational overhead and improves practical deployability.

*4.2 Safety-aware world model learning*

In this section, we consider the augmented constrained MDP $\mathcal{M} = (\mathcal{S}, \mathcal{A}, \mathcal{T}, \mathcal{R}, \mathcal{C}, \mathcal{A}^{\mathcal{S}}, \mu, \gamma)$ under an offline safe RL setting, where the world model “dreams” future rollouts together with safety estimates for policy training. Given a sequence $\mathcal{B} = (x_{1:T}, a_{1:T}, r_{1:T}, c_{1:T}, \alpha_{1:T})$ sampled from the offline dataset, we adopt DreamerV3 (Hafner et al., 2023) as the world model to generate rollouts with explicit safety indicators:

$$\begin{array}{cc}
\text{Observation encoder:} & z_{1:T}^{0} \sim q_{\theta}(z \mid h_{1:T}, x_{1:T}) \\
\text{Dynamics predictor:} & \hat{z}_{1:T}^{l} \sim p_{\theta}(z \mid h_{1:T}^{l}) \\
\text{Sequence model:} & h_{1:T}^{l+1} = f_{\theta}(h_{1:T}^{l}, z_{1:T}^{l}, a_{1:T}) \\
\text{Actor:} & \hat{a}_{1:T}^{l} \sim \pi_{\psi}(\mathrm{a} \mid z_{1:T}^{l}) \\
\text{Decoder:} & \hat{x}_{1:T}^{l} \sim p_{\theta}(x \mid h_{1:T}^{l}, z_{1:T}^{l}) \\
\text{Reward predictor:} & \hat{r}_{1:T}^{l} \sim p_{\theta}(r \mid h_{1:T}^{l}, z_{1:T}^{l}) \\
\text{Cost predictor:} & \hat{c}_{1:T}^{l} \sim p_{\theta}(c \mid h_{1:T}^{l}, z_{1:T}^{l}) \\
\text{Continuation predictor:} & \hat{b}_{1:T}^{l} \sim p_{\theta}(b \mid h_{1:T}^{l}, z_{1:T}^{l}) \\
\text{Safety predictor:} & \hat{\alpha}_{1:T}^{l} \sim p_{\theta}(\alpha \mid h_{1:T}^{l}, z_{1:T}^{l})
\end{array} \tag{8}$$

where $l$ denotes the rollout horizon. The world model receives a trajectory of observations $x_{1:T}$, e.g., the agent's bird's-eye-view (BEV) images. These observations are encoded into latent representations $z_{1:T}^{0} \sim q_\theta(z_{1:T} \mid h_{1:T}, x_{1:T})$, which are then used to generate imagined rollouts. The sequence model then updates the recurrent states according to $h_{1:T}^{l+1} = f_\theta(h_{1:T}^{l}, z_{1:T}^{l}, a_{1:T})$, based on the recurrent states $h_{1:T}^{l}$, hidden presentation $z_{1:T}^{l}$, and action $a_{1:T}$. We perform multi-step prediction at each timestep. Therefore, we use a dynamics predictor to predict the hidden presentation $\hat{z}_{1:T}^{l}$ based solely on $h_{1:T}^{l}$. We also use a reward predictor, a continuation predictor, a cost predictor and a safety predictor to predict the reward, continuity, cost, and safety probability, respectively. Finally, the decoder reconstructs the original observation to ensure that $\{h_{1:T}, z_{1:T}\}$ capture all necessary information.

The following loss functions are used to train the world model:

$$\mathcal{L}_{\text{pred}}(\theta) = -\ln\, p_\theta(x \mid h_{1:T}^{l}, z_{1:T}^{l}) - \ln\, p_\theta(r \mid h_{1:T}^{l}, z_{1:T}^{l}) - \ln\, p_\theta(c \mid h_{1:T}^{l}, z_{1:T}^{l}) - \ln\, p_\theta(\alpha \mid h_{1:T}^{l}, z_{1:T}^{l}) \tag{9}$$

$$\mathcal{L}_{\text{dyn}}(\theta) = \max\big(1, \mathrm{KL}\big[\mathrm{sg}\left(q_\theta(z \mid h_{1:T}, x_{1:T})\right) \parallel\ p_\theta(z \mid h_{1:T}^{l})\big]\big) \tag{10}$$

$$\mathcal{L}_{\text{rep}}(\theta) = \max\left(1, \mathrm{KL}\left[sg\left(p_\theta(z \mid h_{1:T}^{l})\right) \parallel q_\theta(z \mid h_{1:T}, x_{1:T})\right]\right) \tag{11}$$

The losses are scaled and summed for calculating gradients:

$$\mathcal{L}(\theta) \doteq \mathrm{E}_{q_\theta} \textstyle\sum_{t=1}^{T} \left(\beta_{\text{pred}} \mathcal{L}_{\text{pred}}^{t}(\theta) + \beta_{\text{dyn}} \mathcal{L}_{\text{dyn}}^{t}(\theta) + \beta_{\text{rep}} \mathcal{L}_{\text{rep}}^{t}(\theta)\right) \tag{12}$$

where $\beta_{\text{pred}}$, $\beta_{\text{dyn}}$, and $\beta_{\text{rep}}$ are hyperparameters that balance different loss terms. A discussion of the latent consistency induced by KL regularization is provided in Appendix A.1.

*4.3 Safe actor-critic learning*

We use the generated imagined rollouts to train actor-critic networks under an offline safe RL setting. Motivated by expectile-based value estimation, we introduce a separate state-value network $V_\varphi(s)$, which depends only on the state, to mitigate overestimation. The value network is optimized via expectile regression as follows:

$$\mathcal{L}_V(\varphi) = \mathbb{E}_{\tau\sim\mathcal{D},\pi_\psi,p_\theta}[|\kappa - \mathbf{1}(\delta_t < 0)| \cdot \delta_t^2], \delta_t = Q_\phi(\hat{s}_t, \hat{a}_t) - V_\varphi(\hat{s}_t), \tag{13}$$

where $\kappa \in (0,1)$ is the expectile parameter, $(\hat{s}_t, \hat{a}_t)$ are sampled from imagined rollouts generated by the learned world model $p_\theta$, and $Q_\phi(\cdot)$ denotes the critic.

We then optimize the critic using the objective in Eq. (14), which consists of two complementary terms:

$$\mathcal{L}_Q(\phi) = \mathcal{L}_Q^1(\phi) + \mathcal{L}_Q^2(\phi). \tag{14}$$

The first term fits the critic to the rollout-based target $R_t^l$ along imagined trajectories:

$$\mathcal{L}_Q^1(\phi) = \frac{1}{LT} \mathbb{E}_{\tau\sim\mathcal{D},\pi_\psi,p_\theta}\left[\sum_{t=1}^{T} \sum_{l=1}^{L} \left(Q_\phi(\hat{s}_t^l, \hat{a}_t^l) - R_t^l\right)^2\right], \tag{15}$$

where $L$ denotes the imagination horizon and $T$ denotes the rollout length. The rollout target is defined recursively by $R_t^L = V_\varphi(\hat{\boldsymbol{s}}_t^L)$ and $R_t^l = \hat{r}_t^l + \gamma\big((1-\lambda)V_\varphi(\hat{s}_t^l) + \lambda R_t^{l+1}\big)$. The second term is a TD-style regularization term that anchors the critic using the real rewards from the offline dataset:

$$\mathcal{L}_Q^2(\phi) = \frac{1}{T} \mathbb{E}_{\tau\sim\mathcal{D},\pi_\psi}\left[\sum_{t=1}^{T} \left(Q_\phi(s_t, a_t) - \left(r_t + \gamma V_\varphi(s_{t+1})\right)\right)^2\right]. \tag{16}$$

This term is particularly important in early offline training stages when the world model may not generate accurate reward signals; the real rewards provide a stable learning signal and prevent the critic from drifting.

For safe policy learning, we treat the cost signal as a risk surrogate and enforce risk-aware learning through a Lagrangian-constrained formulation under a KL trust region (behavior regularization). Specifically, we optimize the policy by maximizing the Lagrangian objective while constraining the deviation from the behavior policy $\pi_b$:

$$\max_{\pi_\psi} \mathbb{E}_s\left[A^r(s,a) - \lambda_p^k A^c(s,a) - \Phi\big(V_\varphi^c(s), \lambda_p^k, \mu^k\big)\right] \ \text{s.t.}\ D_{\text{KL}}\big(\pi_\psi \| \pi_b\big) \le \epsilon, \tag{17}$$

where $A^r(s,a) = Q_\phi^r(s,a) - V_\varphi^r(s)$ and $A^c(s,a) = Q_\phi^c(s,a) - V_\varphi^c(s)$ are the reward and cost advantages, $\lambda_p^k$ is the Lagrange multiplier and $\mu^k$ is the penalty coefficient, and $\Phi(\cdot)$ is the constraint-handling penalty defined in:

$$\Phi\left(V_{\varphi}^{c}, \lambda_{p}^{k}, \mu^{k}\right)=\begin{cases}\lambda_{p}^{k} V_{\varphi}^{c}+\dfrac{\mu^{k}}{4}\left(V_{\varphi}^{c}\right)^{2}, & \text{if } \lambda_{p}^{k}+\dfrac{\mu^{k}}{2} V_{\varphi}^{c}>0, \\ -\dfrac{\left(\lambda_{p}^{k}\right)^{2}}{\mu^{k}}, & \text{otherwise.}\end{cases} \tag{18}$$

The safety mechanism is realized through the primal-dual (Lagrangian) updates: the actor performs the primal update to improve the objective, while the multiplier $\lambda_p^k$ is updated to enforce the risk constraint (increasing $\lambda_p^k$ when the expected cost exceeds the risk budget and decreasing it otherwise). A typical update takes the form $\lambda_p^{k+1}=\left[\lambda_p^k+\eta_\lambda(\hat{J}_c-d)\right]_+$, where $\hat{J}_c$ is the estimated expected cost and $d$ is the risk budget.

Under the KL constraint, the optimal policy admits an exponential-tilting form (See Appendix A.2 for more details):

$$\pi^*(a \mid s)=\frac{1}{Z(s)} w(s, a) \pi_b(a \mid s), \tag{19}$$

where $Z(s)$ is the normalizing constant and $w(s,a)$ is a nonnegative weighting term that determines how actions from the behavior policy are reweighted during policy improvement. In the next section, we specify how $w(s,a)$ is instantiated using semantic safety guidance.

*4.4 VLM as safety guidance for policy update*

The Lagrangian backbone in Section 4.3 controls risk through the cost advantage and the multiplier updates. We further introduce a semantic safety score $\alpha(s) \in [0,1]$ to modulate how aggressively the policy performs reward-oriented improvement across different contexts. Here $\alpha(s)$ is obtained from the learned safety predictor (trained using the offline VLM labels in Section 4.1), so the policy does not require invoking the VLM online at inference time. Intuitively, when $\alpha(s)$ is high, the state is judged semantically safe and the policy can focus more on improving efficiency; when $\alpha(s)$ is low, reward-driven updates should be conservative and the Lagrangian risk control should dominate.

In our implementation, the action weighting is constructed from the reward advantage. We first define an advantage-based weight:

$$w(s_t, a_t)=\exp\left(\beta A^r(s_t, a_t)\right), \tag{20}$$

where $\beta>0$ controls the sharpness of the advantage weighting.

We then apply a state-level safety gate to obtain the final sample weight:

$$\tilde{w}_t=\left(\epsilon+(1-\epsilon)\alpha(s_t)\right) w(s_t, a_t)=\left(\epsilon+(1-\epsilon)\alpha(s_t)\right)\exp\left(\beta A^r(s_t, a_t)\right), \tag{21}$$

where $\epsilon \in (0,1)$ sets a minimum update strength to avoid vanishing learning in low-safety states. This design makes the reward-driven update strong in semantically safe contexts and automatically suppresses aggressive reward optimization in risky contexts, while risk reduction is still primarily enforced by the Lagrangian terms in Section 4.3. To better understand the role of the semantic gate in Eq. (21), we first summarize several basic properties of the gated weighting scheme.

**Lemma 1 (Bounded and non-vanishing sample weight).** Let $\alpha(s) \in [0,1], \epsilon \in (0,1)$, and $\beta>0$. For the semantic-gated weight:

$$\tilde{w}(s, a)=\left(\epsilon+(1-\epsilon)\alpha(s)\right)\exp\left(\beta A_r(s, a)\right), \tag{22}$$

the following bounds hold for any $(s,a)$:

$$\epsilon \exp\left(\beta A_r(s, a)\right) \leq \tilde{w}(s, a) \leq \exp\left(\beta A_r(s, a)\right). \tag{23}$$

**Proof.** Since $\alpha(s) \in [0,1]$, we have:

$$\epsilon \leq \epsilon+(1-\epsilon)\alpha(s) \leq 1. \tag{24}$$

Multiplying the inequality by the positive term $\exp(\beta A_r(s,a))$ yields the desired result.

**Lemma 2 (Monotonic semantic modulation).** For any fixed state-action pair $(s,a)$, the semantic-gated weight $\tilde{w}(s,a)$ is monotonically increasing in $\alpha(s)$. More precisely,

$$\frac{\partial \tilde{w}(s, a)}{\partial \alpha(s)}=(1-\epsilon)\exp\left(\beta A_r(s, a)\right)>0. \tag{25}$$

**Proof.** Treating $A_r(s,a)$ as fixed for a given $(s,a)$, we differentiate Eq. (22) with respect to $\alpha(s)$. Since $\epsilon \in (0,1)$ and $\exp\left(\beta A_r(s,a)\right)>0$, the derivative is strictly positive.

These properties characterize the semantic gate at the weight level. We next incorporate the gated weight into the behavior-regularized policy objective by minimizing the following weighted negative log-likelihood loss:

$$\mathcal{L}_\pi(\psi) = -\mathbb{E}_{\tau\sim\mathcal{D}}\left[\sum_{t=1}^{T} \tilde{w}_t \log \pi_\psi(a_t \mid s_t)\right] + \eta\mathbb{E}\left[\sum_{t=1}^{T} \mathcal{H}\left(\pi_\psi(\cdot\mid s_t)\right)\right] + \mathbb{E}\left[\sum_{t=1}^{T} \Phi\left(V_\varphi^c(s_t), \lambda_p^k, \mu^k\right)\right], \tag{26}$$

where $\eta$ is an entropy coefficient and $\Phi(\cdot)$ is the same constraint-handling penalty as in Eq. (18). Based on Eq. (26), the next property clarifies how semantic safety affects policy learning. In particular, the semantic gate does not alter the reward-advantage ordering among actions within the same state; instead, it modulates the overall strength of reward-driven policy improvement across states.

**Lemma 3 (Preservation of reward-advantage ordering within each state).** For any fixed state $s$ and any two actions $a_1, a_2$, the semantic gate does not alter the ordering induced by the reward advantage:

$$\tilde{w}(s, a_1) \geq \tilde{w}(s, a_2) \Leftrightarrow A_r(s, a_1) \geq A_r(s, a_2). \tag{27}$$

Equivalently,

$$\frac{\tilde{w}(s, a_1)}{\tilde{w}(s, a_2)} = \exp\left(\beta\left(A_r(s, a_1) - A_r(s, a_2)\right)\right). \tag{28}$$

**Proof.** For a fixed state $s$, the semantic gate,

$$g(s) = \epsilon + (1-\epsilon)\alpha(s) \tag{29}$$

is independent of the action $a$. Therefore,

$$\frac{\tilde{w}(s, a_1)}{\tilde{w}(s, a_2)} = \frac{g(s)\exp\left(\beta A_r(s, a_1)\right)}{g(s)\exp\left(\beta A_r(s, a_2)\right)} = \exp\left(\beta\left(A_r(s, a_1) - A_r(s, a_2)\right)\right). \tag{30}$$

Since the exponential function is strictly increasing, the ordering is preserved.

**Corollary 1 (State-wise attenuation of reward-driven policy update).** Consider the weighted negative log-likelihood term in Eq. (22). For any sample$(s_t, a_t)$, its reward-driven gradient contribution is scaled by $\tilde{w}_t$. Hence, for two states $s_i, s_j$ with the same reward advantage value $A_r$, their relative update magnitudes satisfy:

$$\frac{\tilde{w}(s_i, a)}{\tilde{w}(s_j, a)} = \frac{\epsilon + (1-\epsilon)\alpha(s_i)}{\epsilon + (1-\epsilon)\alpha(s_j)}. \tag{31}$$

In particular, if $\alpha(s_i) < \alpha(s_j)$, then the reward-driven update at $s_i$ is strictly smaller.

**Proof.** This follows directly from Eq. (22) and the definition of $\tilde{w}$.

Overall, Section 4.3 provides a principled primal-dual safe RL backbone that explicitly controls risk (cost), while Section 4.4 injects semantic safety guidance through a state-level gate that adaptively scales the strength of reward-oriented policy improvement.

## 5. EXPERIMENTS

*5.1 Experimental setup*

**Simulation platform.** We use CarDreamer (Gao et al., 2024), an autonomous driving platform built on the CARLA simulator (Dosovitskiy et al., 2017), as our simulation platform under the default settings. It provides a variety of challenging driving tasks. In total, we evaluate our method on ten tasks, including four-lane, lane-merge, navigation, roundabout, left-turn (simple/medium/hard), and right-turn (simple/medium/hard). For clarity of presentation, the results of the three left-turn variants and the three right-turn variants are averaged, and therefore the main comparison tables report six aggregated task categories: four-lane, lane-merge, left-turn, navigation, right-turn, and roundabout.

**Observation space**. We use a bird's-eye-view (BEV) image as the input, where the planned trajectory and the intentions of background vehicles are rendered. The BEV observation has a shape of (128, 128, 3).

**Action space**. The action space contains 15 discrete actions and is composed of all combinations of discrete acceleration and steering values. Specifically, it is defined as the Cartesian product of three discrete acceleration values, [-2.0, 0.0, 2.0], and five discrete steering values, [-0.6, -0.2, 0.0, 0.2, 0.6], resulting in 3 × 5 = 15 discrete actions.

**Data collection**. Our RL framework is mainly trained in an offline setting. Accordingly, we first collect offline data for training. In CarDreamer, we collect offline data using well-trained DreamerV3 agents.

**Network structure**. We use MLPs to process vector inputs and CNNs to process images. When the world model is used, the policy is trained on latent representations instead of raw inputs, following the same architecture as (Hafner et al., 2023).

**Hyperparameters.** World model has 512 GRU recurrent units, 32 CNN multipliers, 1024 dense hidden units, and 5 MLP layers. We use categorical distributions to represent the deterministic and stochastic variables in the RSSM, which provides greater expressivity than Gaussian distributions, as discussed in DreamerV3 (Hafner et al., 2023). The cost limit is set to 10 and the cost weight is set to 10. The length of imagined rollouts is 16. The maximum episode length in CarDreamer is 333 simulation steps, and running out of road is treated as a termination condition. As discussed above, we apply weighted loss terms to balance the various learning objectives during the training of world models. The loss scales for different components are configured as follows: image reconstruction (1.0), vector reconstruction (1.0), reward (1.0), cost (1.0), continuation signal (1.0), dynamics consistency (0.5), representation learning (0.1), actor loss (1.0), critic loss (1.0), target network regularization (1.0), and safety prediction (1.0). Finally, we use 500k training steps for VLM-SAFE.

**Driving scenarios**. All tasks in the CarDreamer simulation platform are conducted on Town 03 and Town 04 maps under CARLA which provides realistic urban and suburban layouts with diverse traffic elements. As shown in Fig. 5, the scenarios cover various road geometries and traffic situations for evaluating policy performance, safety, and robustness in autonomous driving.

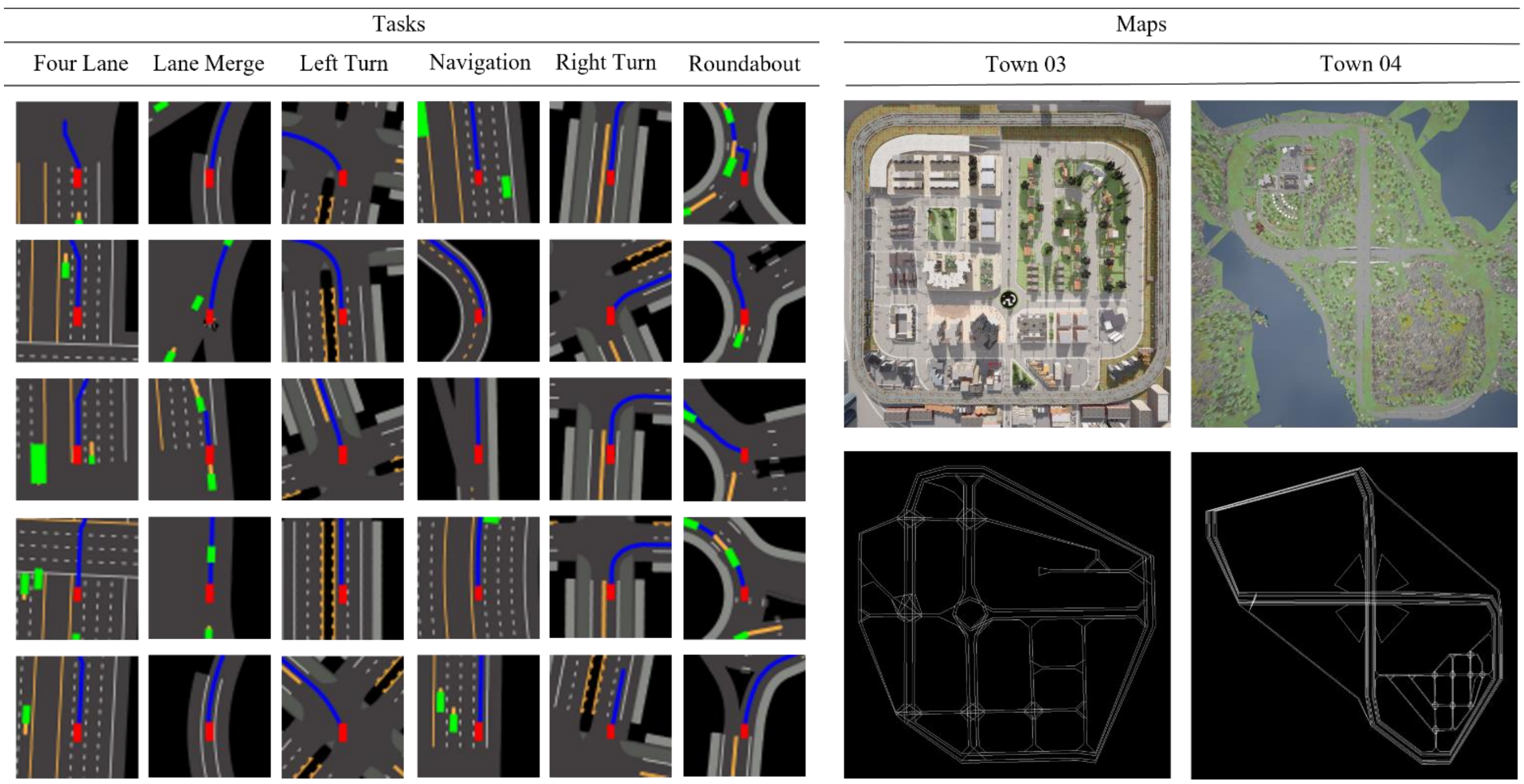

**Fig. 5.** Representative driving tasks and maps in CarDreamer.

*5.2 Evaluation metrics*

We selected three key metrics for evaluation. Driving score is defined as the cumulative reward over one episode. Cost is defined as the cumulative safety cost over one episode, where the step-wise cost is constructed as a weighted combination of collision risk, TTC-based temporal risk, Field-Distance Potential Field (FDPF)-based spatial risk, and lane-departure risk. Arrive rate (AR) is defined as the proportion of successfully completed routes among the latest 10 episodes.

*5.3 Comparison experiments*

We compare VLM-SAFE with Behavioral Cloning (BC), Conservative Q-Learning (CQL) (Kumar et al., 2020), the world-model baseline DreamerV3 (Hafner et al., 2023), and FOSP (Cao et al., 2024) on six driving tasks: four-lane, lane-merge, left-turn, navigation, right-turn, and roundabout. Here, the reported left-turn and right-turn results are not from a single scenario, but are averaged over the corresponding simple, medium, and hard variants provided in CarDreamer. As shown in Tables 1-3, VLM-SAFE does not achieve the best result on every metric in every task, but it provides a more favorable overall safety-performance trade-off across diverse driving scenarios.

In terms of driving score, VLM-SAFE achieves the highest mean performance in five out of the six tasks, namely four-lane, lane-merge, left-turn, right-turn, and roundabout, while DreamerV3 attains the highest score in navigation. This is consistent with the design objective of VLM-SAFE as a safety-oriented RL framework: unlike purely reward-driven baselines, VLM-SAFE explicitly incorporates semantic safety guidance and constrained policy improvement, which may trade some efficiency for safer behavior in reward-dominant scenarios such as navigation.

From the safety perspective, VLM-SAFE achieves the lowest mean cost in four-lane, lane-merge, and roundabout, while remaining competitive in the other tasks. Although some baselines obtain lower costs in specific scenarios, these gains are often accompanied by lower driving scores or weaker completion performance. The arrive rate results are more mixed: VLM-SAFE achieves the best arrive rate in lane-merge and left-turn, and is the only method with a nonzero arrive rate in navigation, but it is not the best in every task. Overall, these results suggest that VLM-SAFE provides a favorable balance between task performance and safety, especially in interactive and safety-critical driving settings.

Table 1
Comparison results on driving score

| Model | Four Lane | Lane Merge | Left Turn | Navigation | Right Turn | Roundabout |
|---|---|---|---|---|---|---|
| BC | 67.83 ± 53.06 | -90.15 ± 97.24 | 101.98 ± 21.36 | 238.32 ± 74.42 | 76.33 ± 27.15 | -137.67 ± 17.78 |
| CQL | 245.87 ± 19.87 | -86.43 ± 68.63 | 112.76 ± 23.54 | 405.36 ± 82.75 | 87.91 ± 31.79 | -58.26 ± 45.49 |
| DreamerV3 | 249.13 ± 181.24 | -30.16 ± 103.75 | 121.20 ± 93.45 | **533.52 ± 288.05** | 13.78 ± 38.86 | -9.83 ± 43.60 |
| FOSP | 159.49 ± 138.04 | 131.22 ± 114.21 | 231.09 ± 28.81 | 373.62 ± 70.43 | 108.03 ± 31.17 | 70.83 ± 21.25 |
| VLM-SAFE (ours) | **271.65 ± 139.81** | **147.03 ± 67.40** | **231.94 ± 26.14** | 492.42 ± 231.95 | **113.08 ± 50.20** | **109.94 ± 16.72** |

Table 2
Comparison results on cost

| Model | Four Lane | Lane Merge | Left Turn | Navigation | Right Turn | Roundabout |
|---|---|---|---|---|---|---|
| BC | 19.52 ± 0.57 | 3.73 ± 0.27 | 1.99 ± 0.17 | **39.43 ± 1.60** | 2.17 ± 0.16 | 4.83 ± 0.30 |
| CQL | 24.35 ± 2.74 | 3.15 ± 0.12 | **1.58 ± 0.09** | 54.29 ± 1.14 | **2.00 ± 0.75** | 5.69 ± 0.67 |
| DreamerV3 | 22.15 ± 6.26 | 2.53 ± 0.32 | 1.74 ± 0.37 | 51.62 ± 1.73 | 2.71 ± 2.28 | 5.77 ± 2.02 |
| FOSP | 19.80 ± 1.52 | 2.61 ± 0.83 | 1.85 ± 0.40 | 45.74 ± 6.58 | 2.58 ± 0.72 | 4.57 ± 2.96 |
| VLM-SAFE (ours) | **18.75 ± 2.77** | **2.48 ± 0.37** | 1.76 ± 0.25 | 42.12 ± 4.13 | 2.78 ± 1.86 | **3.65 ± 0.05** |

Table 3
Comparison results on arrive rate

| Model | Four Lane | Lane Merge | Left Turn | Navigation | Right Turn | Roundabout |
|---|---|---|---|---|---|---|
| BC | 0.19 ± 0.05 | 0.09 ± 0.05 | 0.35 ± 0.03 | 0.00 ± 0.00 | 0.41 ± 0.03 | 0.02 ± 0.01 |
| CQL | **0.59 ± 0.06** | 0.08 ± 0.05 | 0.51 ± 0.05 | 0.00 ± 0.00 | **0.60 ± 0.04** | 0.08 ± 0.03 |
| DreamerV3 | 0.19 ± 0.06 | 0.10 ± 0.08 | 0.51 ± 0.10 | 0.00 ± 0.00 | 0.20 ± 0.21 | **0.10 ± 0.09** |
| FOSP | 0.10 ± 0.11 | 0.15 ± 0.15 | 0.37 ± 0.20 | 0.00 ± 0.00 | 0.10 ± 0.16 | 0.04 ± 0.02 |
| VLM-SAFE (ours) | 0.20 ± 0.09 | **0.26 ± 0.10** | **0.52 ± 0.07** | **0.01 ± 0.00** | 0.19 ± 0.24 | 0.08 ± 0.03 |

### *5.4 Ablation experiments*

We first analyze the effect of incorporating the Lagrangian safe RL mechanism into the world model. Compared with the DreamerV3 baseline, the Safe RL variant substantially improves driving score in several tasks, especially in navigation, indicating that explicit cost-aware optimization can encourage more active task completion instead of overly passive behavior. However, this gain does not uniformly translate into safer behavior, since the resulting costs remain high or even increase in several scenarios, suggesting that static constraint handling alone is insufficient to achieve a desirable safety–performance balance.

Table 4
Ablation results on driving score

| Model | Four Lane | Lane Merge | Left Turn | Navigation | Right Turn | Roundabout |
|---|---|---|---|---|---|---|
| Baseline (DreamerV3) | 249.13 ± 181.24 | -30.16 ± 103.75 | 121.20 ± 93.45 | 533.52 ± 288.05 | 13.78 ± 38.86 | -9.83 ± 43.60 |
| + safe RL | **283.06 ± 29.66** | 78.51 ± 18.68 | 215.40 ± 79.82 | **603.29 ± 152.40** | 60.93 ± 64.05 | 103.61 ± 37.95 |
| + VLM-based safety guidance | 271.65 ± 139.81 | **147.03 ± 67.40** | **231.94 ± 26.14** | 492.42 ± 231.95 | **113.08 ± 50.20** | **109.94 ± 16.72** |

Table 5
Ablation results on cost

| Model | Four Lane | Lane Merge | Left Turn | Navigation | Right Turn | Roundabout |
|---|---|---|---|---|---|---|
| Baselines (DreamerV3) | 22.15 ± 6.26 | 2.53 ± 0.32 | **1.74 ± 0.37** | 51.62 ± 1.73 | **2.71 ± 2.28** | 5.77 ± 2.02 |
| + safe RL | 21.59 ± 3.54 | 2.64 ± 0.04 | 3.81 ± 3.23 | 46.39 ± 0.11 | 3.67 ± 2.58 | 4.09 ± 0.10 |
| + VLM-based safety guidance | **18.75 ± 2.77** | **2.48 ± 0.37** | 1.76 ± 0.25 | **42.12 ± 4.13** | 2.78 ± 1.86 | **3.65 ± 0.05** |

We then evaluate the full model by further introducing VLM-based semantic safety guidance. Compared with the Safe RL variant, the full VLM-SAFE model consistently reduces cost across all reported tasks, while maintaining competitive driving performance. For example, in the four-lane scenario, the cost decreases from 21.59 to 18.75, and in navigation it decreases from 46.39 to 42.12. Although this safety-oriented adjustment may reduce raw score in some efficiency-dominant settings, it leads to a more favorable trade-off between task performance and safety, which is the primary design objective of VLM-SAFE.

### *5.5. Analysis of VLM safety scoring*

In this section, we analyze the capability of the VLM-based safety scoring mechanism to distinguish between safe and dangerous driving situations. As shown in Fig. 6, the predicted VLM safety score exhibits a generally decreasing trend as the environmental cost increases. This suggests that, under standard driving situations, the semantic safety signal produced by the VLM is broadly aligned with the underlying risk level: lower-risk states tend to receive higher safety scores, whereas riskier states tend to receive lower ones. The successful examples in Fig. 7 further show that the model captures useful semantic cues in typical traffic scenes and can distinguish many common safe and unsafe situations.

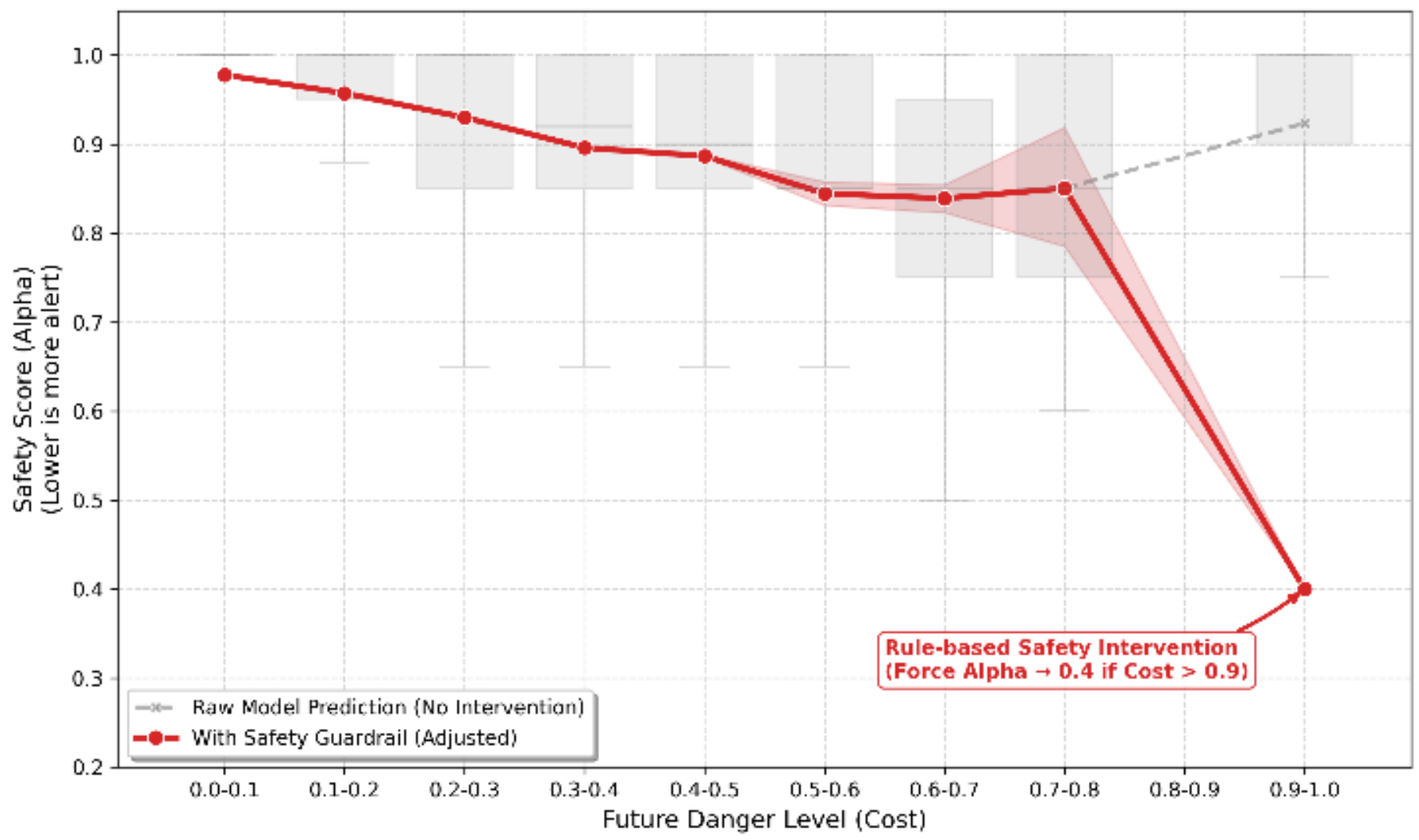

**Fig. 6.** Relationship between environmental-cost bins and predicted VLM safety scores. The score generally decreases as cost increases, while a conservative rule-based correction is applied for rare extreme-risk states.

However, this tendency is not perfect in rare extreme edge cases. In particular, when the environmental cost exceeds 0.9, the corresponding state usually indicates that a collision is imminent or has already occurred. In this regime, the raw VLM safety score may remain unexpectedly high rather than continuing to decrease. We attribute this anomaly to two main factors. First, current pretrained VLMs (we use Qwen-VL-8B (Bai et al., 2025) in our experiments) still have inherent capability limitations, especially for zero-shot accident recognition in complex simulated driving environments. Second, our offline dataset is collected mainly using expert agents, which rarely trigger collisions actively. As a result, the few samples in the highest-cost bin are often atypical accident cases, such as rear-end collisions caused by another vehicle rather than by the ego vehicle's own unsafe action. Since the VLM only receives front-view RGB images together with structured prompts, it cannot reliably access rear-view context. Therefore, some collision events are only weakly observable from the current input, making such failure cases understandable under the present sensing setup. Expanding the visual input to include multi-camera panoramic views would be a promising direction for future improvement.

To improve robustness under these rare edge-case failures, we introduce a conservative safeguard during offline label generation. Specifically, when the environmental cost exceeds 0.9, we cap the VLM safety score at 0.4, corresponding to the "danger" level in Fig. 4. This rule-based correction is designed only for rare extreme-risk states and serves as a conservative offline label adjustment rather than as a core component of the online policy learning pipeline. Its purpose is to prevent the learning process from being affected by overly optimistic safety labels in actual collision states and to provide a more stable supervision signal for subsequent policy learning.

Importantly, this limitation does not negate the usefulness of the proposed semantic safety scoring mechanism. Although the VLM is imperfect in extreme edge cases, the overall trend in Fig. 6 still demonstrates that the predicted safety score is meaningfully correlated with the environmental risk level in most standard situations. More importantly, the ablation results in Tables 4 and 5 show that removing semantic safety guidance leads to a worse overall safety-performance trade-off. Therefore, while the VLM-based safety signal is not flawless, it still provides useful semantic supervision that improves policy learning in practice.

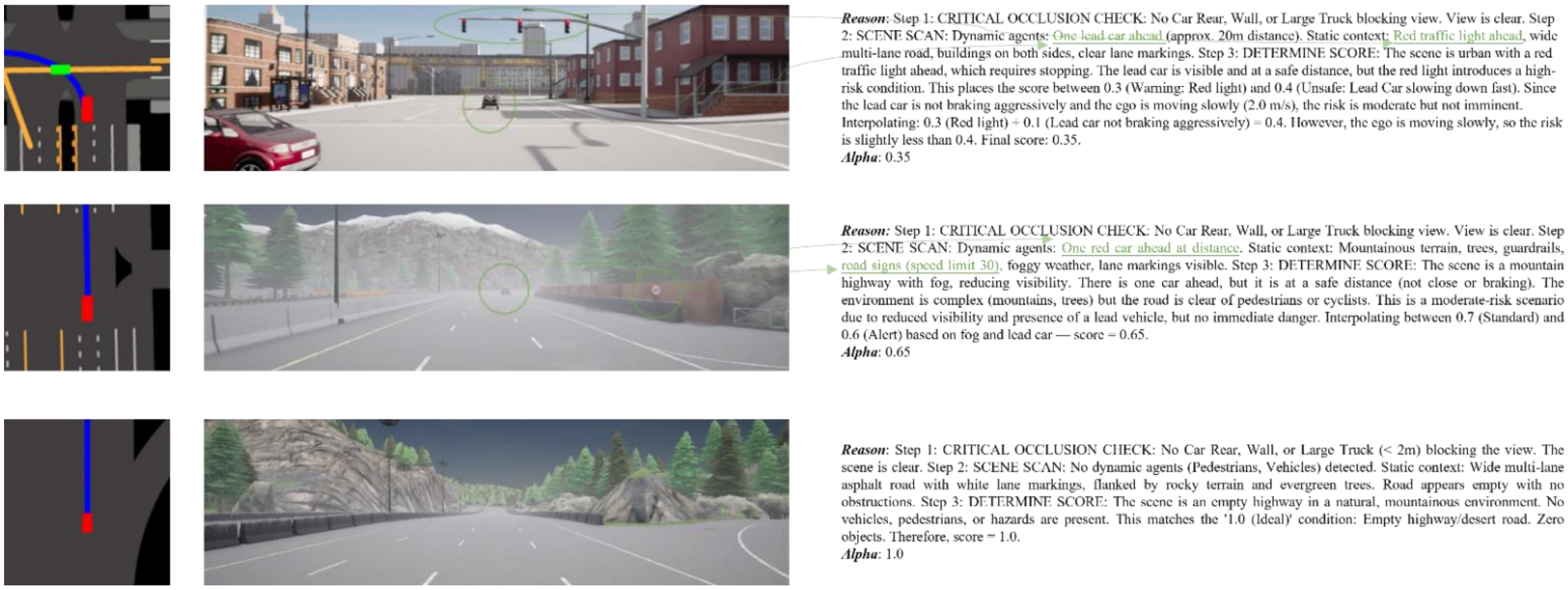

**Fig. 7.** Examples of accurate VLM safety estimation.

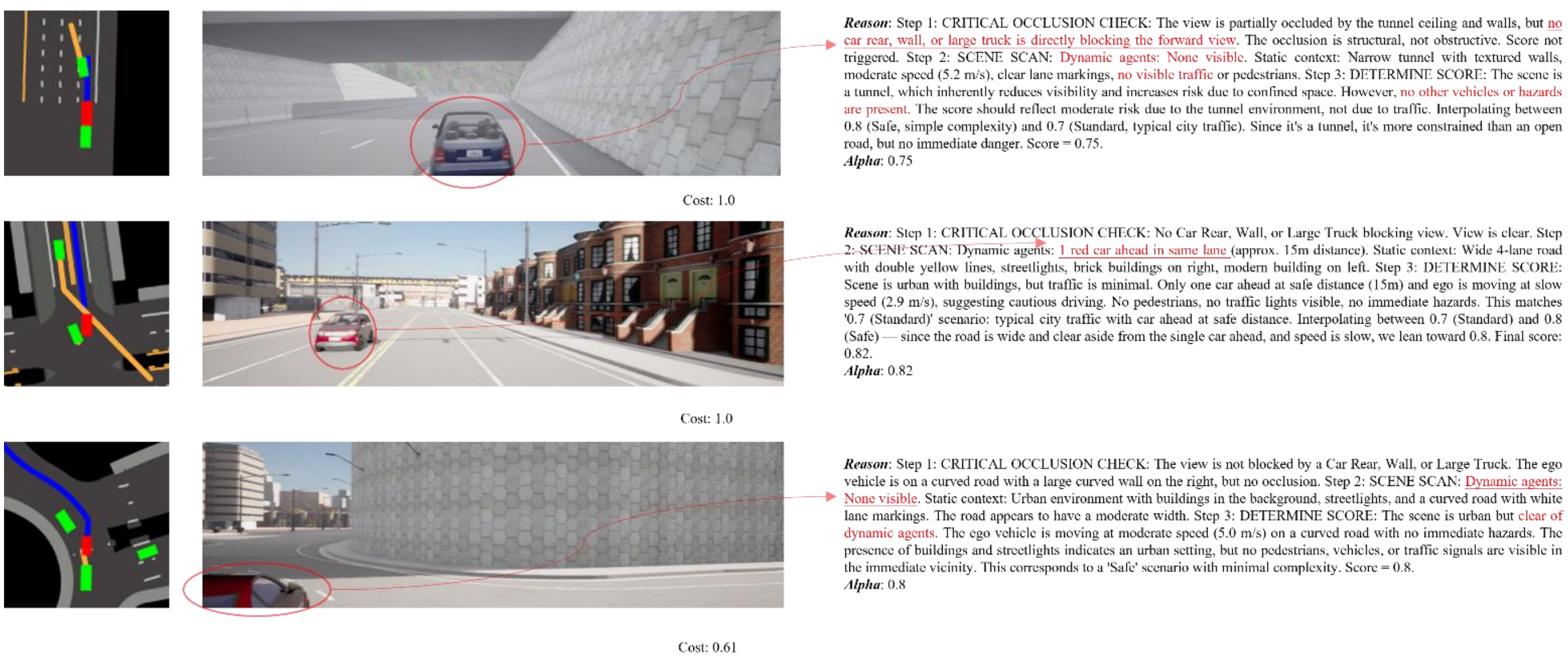

**Fig. 8.** Examples of failed or inaccurate VLM safety estimation.

### *5.6. Analysis of traffic-density shift*

To further evaluate robustness under distribution shift within the same scenario family, we construct a traffic-density-shift benchmark in the four-lane scenario by increasing the number of background vehicles during evaluation. As shown in Table 6, denser traffic consistently degrades the performance of both methods, confirming that higher traffic density substantially increases task difficulty. For FOSP, the driving score drops from 159.49 to 111.16 and 65.86, corresponding to relative decreases of 30.3% and 58.7%, respectively. In comparison, the driving score of VLM-SAFE decreases from 271.65 to 196.10 and 140.04, corresponding to smaller relative drops of 27.8% and 48.4%. This indicates that VLM-SAFE preserves driving performance better as interaction density increases.

Moreover, VLM-SAFE consistently outperforms FOSP in absolute performance across all density levels. Specifically, compared with FOSP, VLM-SAFE achieves 70.3% higher driving score at 50 vehicles, 76.4% higher at 100 vehicles, and 112.6% higher at 150 vehicles. The widening score margin under denser traffic suggests that the proposed semantic safety guidance becomes more beneficial when decision-making must be carried out under more complex and safety-critical interactions.

A similar trend can be observed in the success rate. FOSP decreases from 0.10 to 0.09 and 0.08 as traffic density increases, while VLM-SAFE decreases from 0.20 to 0.19 and 0.15. Although both methods experience reduced completion under denser traffic, VLM-SAFE maintains consistently higher success rates than FOSP at all density levels, with relative improvements of 100.0%, 111.1%, and 87.5%, respectively. Overall, these results suggest that VLM-SAFE maintains better mean performance than FOSP under traffic-density shift, while its driving score degrades more slowly as traffic density increases.

Table 6
Performance under traffic-density Shift in the four-Lane Scenario

| No. of Vehicles | Model | Driving Score | Success Rate |
|---|---|---|---|
| 50 | FOSP | 159.49 ± 138.04 | 0.10 ± 0.11 |
| | VLM-SAFE | **271.65 ± 139.81** | 0.20 ± 0.09 |
| 100 | **FOSP** | 111.16 ± 106.54 | 0.09 ± 0.09 |
| | **VLM-SAFE** | 196.10 ± 128.48 | 0.19 ± 0.11 |
| 150 | FOSP | **65.86 ± 72.02** | **0.08 ± 0.06** |
| | VLM-SAFE | **140.04 ± 116.46** | **0.15 ± 0.07** |

*5.7. Analysis of the imagination capability of the world model*

As shown in Fig. 9, we compare ground-truth observations with predictions generated by the world model. Specifically, the top row presents the actual observed frames, the middle row shows the predicted future frames by the world model, and the bottom row highlights the discrepancies between the two.

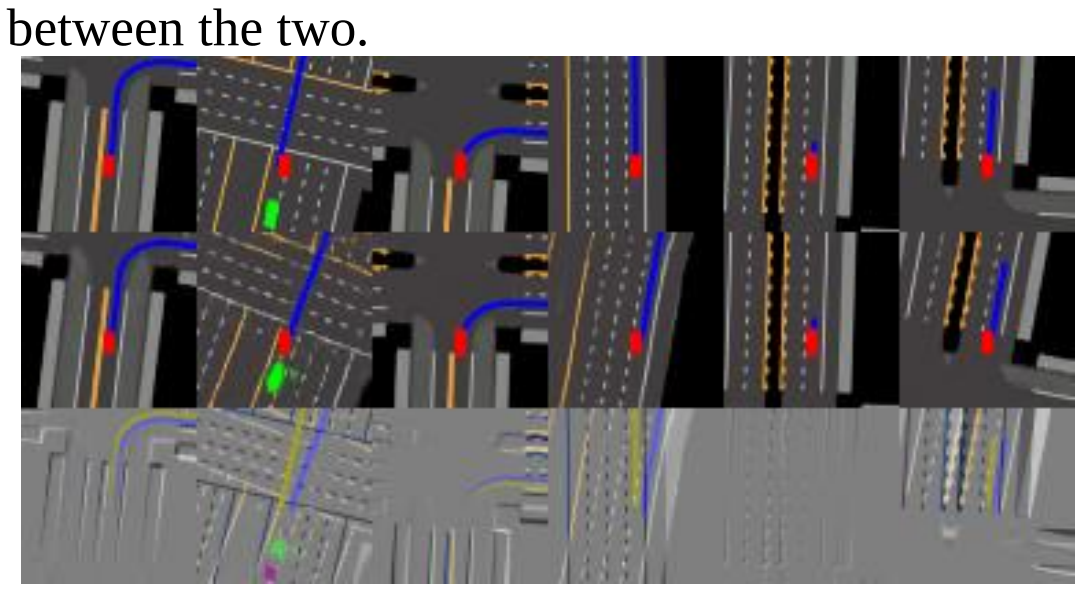

**Fig. 9.** Comparison between ground-truth observations and world-model predictions.

The results suggest that the learned world model can reasonably anticipate short-horizon future scenes within the CarDreamer simulation environment. In particular, it captures major vehicle motions, including straight driving, right turns, and the ego-relative rotation and translation of the BEV representation.

## 6. Conclusions and future work

This paper presented VLM-SAFE, a novel framework for safe autonomous driving policy learning that addresses key limitations of RL-based methods, including low sample efficiency, limited robustness under distribution shift, and unsafe online trial-and-error interaction. More importantly, VLM-SAFE is built upon a human cognitive loop of observe-imagine-evaluate-act: it observes offline driving data and scene context, imagines future trajectories through a learned world model, evaluates these imagined outcomes with VLM-grounded semantic safety guidance, and finally improves the policy via safety-aware actor-critic learning. By tightly coupling perception, future prediction, semantic safety assessment, and decision-making within a unified closed loop, VLM-SAFE enables more effective and semantically grounded safe policy learning. Extensive simulation experiments suggest that VLM-SAFE improves safety and robustness while achieving a more favorable safety–performance trade-off than representative baselines. In particular, the proposed semantic safety guidance is especially beneficial in dense and safety-critical interaction settings.

Nevertheless, this study still has limitations. All experiments were conducted in simulation, where rule-based background vehicles cannot fully capture the diversity and stochasticity of real-world traffic, and the pretrained VLMs used for safety guidance may still lack domain-specific understanding of driving scenarios. These factors may limit the realism of the imagine step and the accuracy of the evaluate step in the cognitive loop. Future work will therefore focus on incorporating more human-like traffic behavior learned from real-world data, developing stronger driving-oriented multimodal safety models, and conducting experiments and validation on real-world platforms (e.g., Sky-Drive (Z. Huang et al., 2025)) to further improve robustness and real-world applicability.

## Appendix

A1. Latent Consistency Induced by KL Regularization

**Remark 1**. Let $z_t \sim q_\phi(z_t \mid h_t, x_t)$ denote the posterior latent inferred from observations, and let $\hat{z}_t \sim p_\phi(z_t \mid h_t)$ denote the prior latent predicted by the latent dynamics model. The KL-based dynamics and representation losses in Eqs. (10) and (11) encourage consistency between the inferred posterior and the predicted prior in latent space.

More specifically, minimizing

$$D_{\mathrm{KL}}\left(q_\phi(z_t \mid h_t, x_t) \| p_\phi(z_t \mid h_t)\right) \tag{32}$$

reduces the mismatch between posterior and prior latent distributions. As a result, the learned world model is encouraged to produce predicted latent states that remain close to the latents inferred from observations, which helps stabilize imagined rollouts and improves short-horizon latent prediction.

**Discussion**. We do not claim a uniform per-step upper bound on $\|\hat{z}_t - z_t\|^2$ here. Instead, the role of KL regularization is interpreted as a stabilizing mechanism that encourages posterior-prior latent consistency during world-model training. This regularization is applied at each training step and empirically helps the latent dynamics remain well aligned with observation-conditioned latent inference.

A2. Extraction of the optimal policy

The following result, adapted from (Nair et al., 2020), characterizes the optimal policy under a KL-constrained objective.

**Proposition 2**. Suppose that the optimizing problem has the following form:

$$\max_{\pi} \mathbb{E}_{a\sim\pi}[A(s,a)] \text{ s.t. } D_{\mathrm{KL}}(\pi_\psi \parallel \pi_b) \leqslant \epsilon \tag{33}$$

The optimal solution will satisfy:

$$\pi^*(a \mid s) \propto \exp\left(\alpha A(s,a)\right)\pi_b(a \mid s) \tag{34}$$

**Proof**. The Lagrangian can be written as:

$$L(\pi_\psi, \lambda, \mu) = \mathbb{E}_{\boldsymbol{a}\sim\pi}[A(\boldsymbol{s},\boldsymbol{a})] - \lambda\left(D_{\mathrm{KL}}(\pi_\psi \parallel \pi_b) - \epsilon\right) \tag{35}$$

Then we take the partial derivative with respect to $\pi$ and set it to 0:

$$\frac{\partial L}{\partial \pi_\psi} = A(\boldsymbol{s},\boldsymbol{a}) - \lambda \log\, \pi_b(\boldsymbol{a} \mid \boldsymbol{s}) + \lambda \log\, \pi_\psi(\boldsymbol{a} \mid \boldsymbol{s}) = 0 \tag{36}$$

We have:

$$\pi^*(a \mid s) = \frac{1}{Z}\exp\left(\alpha A(s,a)\right)\pi_b(a \mid s) \tag{37}$$

## Acknowledgement

Funding for this research was provided by the Center for Connected and Automated Transportation under Grant No. 69A3552348305 of the U.S. Department of Transportation, Office of the Assistant Secretary for Research and Technology (OST-R), University Transportation Centers Program. General support was provided by the Center for Innovation in Control, Optimization, and Networks (ICON) and the Autonomous and Connected Systems (ACS) initiatives at Purdue University's College of Engineering. In addition, this research was supported by grants from NVIDIA and utilized NVIDIA RTX PRO 6000 Blackwell Max-Q Workstation Edition GPUs and A100 GPU-Hours on Saturn Cloud.